\documentclass[acmsmall,screen]{acmart}
\AtBeginDocument{%
  }

\setcopyright{acmcopyright}
\copyrightyear{2023}
\acmYear{2023}
\acmDOI{XXXXXXX.XXXXXXX}

\acmJournal{JACM}
\acmVolume{37}
\acmNumber{4}
\acmArticle{111}
\acmMonth{8}

\usepackage{graphicx}
\usepackage{amsmath}
\usepackage{amsthm}
\usepackage{algorithm}
\usepackage{algorithmic}
\usepackage{subfigure}
\usepackage{booktabs}

\usepackage{bm}

\usepackage{amssymb}
\usepackage{latexsym}
\usepackage{dsfont}
\usepackage{multirow}
\usepackage{color}

\newcommand{\tabincell}[2]{\begin{tabular}{@{}#1@{}}#2\end{tabular}}

\begin{document}

\title{Transform-Equivariant Consistency Learning for Temporal Sentence Grounding}

\author{Daizong Liu}
\affiliation{%
  \institution{Huazhong University of Science and Technology}
  \country{China}
}

\author{Xiaoye Qu}
\affiliation{%
  \institution{Huazhong University of Science and Technology}
  \country{China}}

\author{Jianfeng Dong}
\affiliation{%
  \institution{Zhejiang Gongshang University}
  \country{China}
}

\author{Pan Zhou}
\affiliation{%
  \institution{Huazhong University of Science and Technology}
  \country{China}}

\author{Zichuan Xu}
\affiliation{%
  \institution{Dalian University of Technology}
  \country{China}}

\author{Haozhao Wang}
\affiliation{%
  \institution{Huazhong University of Science and Technology}
  \country{China}}

\author{Xing Di}
\affiliation{%
  \institution{Protagolabs Inc.}
  \country{USA}}

\author{Weining Lu}
\affiliation{%
  \institution{Tsinghua University}
  \country{China}}

\author{Yu Cheng}
\affiliation{%
  \institution{Microsoft Research}
  \country{USA}}

\renewcommand{\shortauthors}{Liu et al.}

\begin{abstract}
This paper addresses the temporal sentence grounding (TSG).
Although existing methods have made decent achievements in this task, they not only severely rely on abundant video-query paired data for training, but also easily fail into the dataset distribution bias. 
To alleviate these limitations, we introduce a novel Equivariant Consistency Regulation Learning (ECRL) framework to learn more discriminative query-related frame-wise representations for each video, in a self-supervised manner. Our motivation comes from that the temporal boundary of the query-guided activity should be consistently predicted under various video-level transformations.
Concretely, we first design a series of spatio-temporal augmentations on both foreground and background video segments to generate a set of synthetic video samples. 
In particular, we devise a self-refine module to enhance the completeness and smoothness of the augmented video.
Then, we present a novel self-supervised consistency loss (SSCL) applied on the original and augmented videos to capture their invariant query-related semantic by minimizing the KL-divergence between the sequence similarity of two videos and a prior Gaussian distribution of timestamp distance. 
At last, a shared grounding head is introduced to predict the transform-equivariant query-guided segment boundaries for both the original and augmented videos.
Extensive experiments on three challenging datasets (ActivityNet, TACoS, and Charades-STA) demonstrate both effectiveness and efficiency of our proposed ECRL framework.
\end{abstract}



\maketitle

\section{Introduction}
\begin{figure}[t]
\centering
\includegraphics[width=0.7\textwidth]{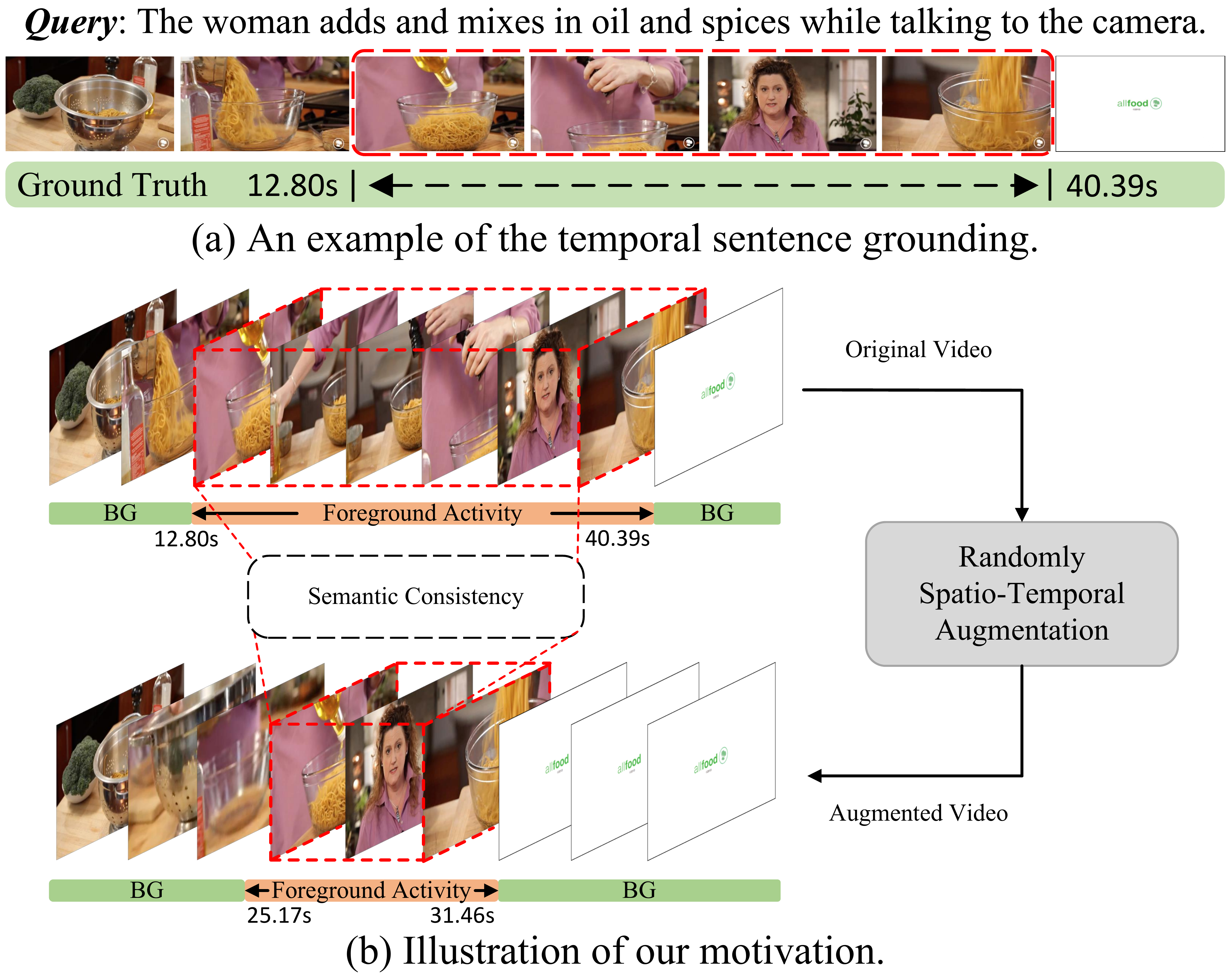}
\caption{(a) An illustrative example of TSG. (b) Illustration of our motivation. ``BG" means the query-irrelevant background. Here, we learn to predict the transform-equivariant temporal boundaries of the query-related segments in both original and augmented videos.}
\label{fig:intro}
\end{figure}

Temporal sentence grounding (TSG) is an important yet challenging task in video understanding \cite{liu2022rethinking,liu2021spatiotemporal,xu2019mhp,liu2020saanet,liu2021video,liu2021f2net}, which has drawn increasing attention over the last few years due to its vast potential applications in video captioning \cite{jiang2018recurrent,chen2020learning,dong2016early}, video summarization \cite{song2015tvsum,chu2015video,yuan2014memory}, video-text retrieval~\cite{dong2021dual,yang2020tree,dong2018predicting}, and video question answering \cite{gao2019structured,le2020hierarchical,qiao2018exploring}, etc. As shown in Figure \ref{fig:intro} (a), this task aims to ground the most relevant video segment according to a given sentence query. It is substantially more challenging as it needs to not only model the complex multi-modal interactions among video and query features, but also capture complicated context information for predicting the accurate query-guided segment boundaries.

Most previous works either follow a proposal-based framework \cite{anne2017localizing,chen2018temporally,zhang2019cross,yuan2019semantic,zhang2019learning,liu2021context,zeng2020dense,zhang2021multi,lan2022a,lan2023a,zeng2022moment,liu2020jointly,liu2022unsupervised,liu2021progressively,liu2021adaptive} that first generates multiple segment proposals and then selects the most query-matched one, or follow a proposal-free framework \cite{chenrethinking,yuan2019find,mun2020local,zhang2020span} that directly regresses the start and end timestamps of the segment with the multi-modal representations. Although they have achieved significant performance, these methods are data-hungry and require a large amount of annotated data for training.
Moreover, recent studies \cite{otani2020uncovering,yuan2021closer,zhang2021towards} point out that existing works rely on exploiting the statistical regularities of annotation distribution for segment prediction, thus easily stucking into the substantial distribution bias existed in benchmark datasets.
Therefore, how to synthesize positive video-query data pairs without human labors while mitigating the data distribution bias during the model training is an emerging issue for TSG task.

In this paper, we propose a novel Equivariant Consistency Regulation Learning (ECRL) framework to address the above issues in a self-supervised manner.
As shown in Figure~\ref{fig:intro} (b), given an untrimmed video that contains a specific segment semantically corresponding to a query, the target segment boundaries would drastically change when applying transformations to the raw video (\textit{e.g.}, random upsampling or downsampling on two background sub-videos and one foreground segment, respectively). This operation does not disrupt the main contents of the video, since the query-aware semantic of target segment is invariant to these transforms. Therefore, synthesizing a set of such augmented videos is more representative than the truncated ones \cite{zhang2021towards}, and can be utilized to assist the model training. Moreover, the stability and generalization ability of the model are crucial for precisely de-limiting the segment boundaries due to the existence of the data distribution bias. To make our ECRL be strongly equivariant with respect to numerous transforms, we further capture the consistent query-related semantics between the original and augmented videos for better discriminating foreground-background frame-wise representations.

To this end, we first introduce a spatio-temporal transformation strategy to apply various video-level augmentations on each video to synthesize new video samples. Considering that general transformations may destroy the continuity of the adjacent frames, we further propose a self-refine module to enhance the completeness and smoothness of the augmented video. Then, we present a novel self-supervised consistency loss (SSCL), which optimizes the frame-wise representations by minimizing the KL-divergence between the sequence similarity of original/augmented videos and a prior Gaussian distribution for capturing the consistent query-related semantics between two videos. At last, a shared grounding head is utilized to predict the transform-equivariant query-guided segment boundaries. In this manner, our ECRL not only can be well-trained with the enriched data samples but also is robust to the data distribution bias.

The main contributions of this work are three-fold:
\begin{itemize}
    \item To our best knowledge, this paper represents the first attempt to explore transform-equivariance for TSG task. Specifically, we propose the novel Equivariant Consistency Regulation Learning (ECRL) framework, to capture the consistency knowledge between the original video and its spatio-temporal augmented variant.
    \item We propose a self-refine module to smooth the discrete adjacent frames of augmented video. Besides, we propose a self-supervised consistency loss (SSCL) to utilize KL-divergence with a prior Gaussian distribution to discriminate frame-wise representation for learning the invariant query-related visual semantic.
    \item Comprehensive evaluations are conducted on three challenging TSG benchmarks: ActivityNet, TACoS, and Charades-STA. Our method re-calibrates the state-of-the-art performance by large margins.
\end{itemize}

\section{Related Work}
\noindent \textbf{Temporal action localization.}
Temporal action localization is a task that involves classifying action instances by predicting their start and end timestamps along with their respective action category labels. This is a single-modal task that has been extensively studied in the literature \cite{sun2021exploiting,pramono2021spatial,zhai2021action}. Researchers have proposed two main categories of methods for temporal action localization, namely one-stage and two-stage methods \cite{lin2017single, xu2019two}.
One-stage methods predict both the action boundaries and labels simultaneously. For instance, Xu \textit{et al.} \cite{xu2020g} used a graph convolutional network to perform one-stage action localization. On the other hand, two-stage methods first generate action proposals and then refine and classify confident proposals. Usually, the confident proposals are generated using the anchor mechanism \cite{xu2019two, yang2020revisiting}. However, there are other methods for generating proposals, such as sliding window \cite{shou2016temporal}, temporal actionness grouping \cite{zhao2017temporal}, and combining confident starting and ending frames \cite{lin2019bmn}.

\noindent \textbf{Temporal sentence grounding.}
Temporal sentence grounding (TSG) is a multimedia task that aims to semantically link a given sentence query with a specific video segment by identifying its temporal boundary. This task was introduced by \cite{gao2017tall} and \cite{anne2017localizing}. TSG is considerably more challenging than temporal action localization, as it requires capturing both visual and textual information and modeling the complex multi-modal interactions between them to accurately identify the target activity.
Unlike temporal action localization, TSG involves identifying the semantic meaning of a sentence query and mapping it to a specific video segment. This requires understanding the context and meaning of the query and interpreting it in relation to the visual content of the video. Additionally, TSG needs to consider the complex interactions between the visual and textual modalities to accurately model the target activity.
Various TSG algorithms  \cite{ge2019mac,chen2018temporally,wang2019temporally,zhang2019man,yuan2019semantic,zhang2019cross,cao2021pursuit,ma2021hierarchical,hu2021coarse,hu2021video,zheng2023progressive,liu2020reasoning,liu2022exploring,liu2022reducing,liu2023exploring,liu2022skimming,fang2022hierarchical,liu2022few,fang2022multi,liu2022learning,fang2023you,xiong2022gaussian,guo2022hybird,liu2023jointly,zhu2023rethinking,xiong2023tracking} have been proposed within the proposal-based framework, which first generates multiple segment proposals, and then ranks them according to the similarity between proposals and the query to select the best matching one. 
Traditional methods for temporal sentence grounding, such as \cite{liu2018attentive} and \cite{gao2017tall}, use video segment proposals to localize the target segment. These methods first sample candidate segments from a video and then integrate the query with segment representations using a matrix operation. However, these methods lack a comprehensive structure for effectively modeling multi-modal feature interactions.
To address this limitation and more effectively mine cross-modal interactions, recent works such as \cite{xu2019multilevel}, \cite{chen2019semantic}, \cite{ge2019mac}, and \cite{zhang2019learninga} have proposed integrating the sentence representation with each video segment individually and then evaluating their matching relationships. By incorporating more fine-grained features from both the visual and textual modalities, these methods can better capture the complex interactions between them and achieve improved performance on TSG tasks.
Although these methods achieve good performances, they severely rely on the quality of the proposals and are time-consuming.
Without using proposals, recent works \cite{rodriguez2020proposal,yuan2019find,chenrethinking,nan2021interventional,mun2020local,zhang2020span,mo2022multi,liu2023hypotheses} directly regress the temporal locations of the target segment. They do not rely on the segment proposals and directly select the starting and ending frames by leveraging cross-modal interactions between video and query. Specifically, they either regress the start/end timestamps based on the entire video representation or predict at each frame to determine whether this frame is a start or end boundary.
However, recent studies \cite{otani2020uncovering,yuan2021closer,zhang2021towards} point out that both types of above works are limited by the issue of distribution bias in TSG datasets and models. In this paper, we propose a novel framework to alleviate the data bias in a self-supervised learning manner.

\noindent \textbf{Self-supervised learning.}
Self supervised learning (SSL) has become an increasingly popular research area in recent years \cite{doersch2015unsupervised,zhang2016colorful,gidaris2018unsupervised,noroozi2017representation,dong2022hierarchical}. In the context of videos, SSL methods have focused on tasks such as inferring the future \cite{han2019video}, discriminating shuffled frames \cite{misra2016shuffle}, and predicting speed \cite{benaim2020speednet}. Some recent works \cite{feichtenhofer2021large,kuang2021video,qian2021spatiotemporal,yao2021seco} have also used contrastive loss for video representation learning, where different frames in a video or different frames in other videos are treated as negative samples.
Different from these methods, our goal is fine-grained temporal understanding of videos and we treat a long sequence of frames as input data. Moreover, since the neighboring frames in video have high semantic similarities, directly regarding these frames as negatives like above works may hurt the learning. To avoid this issue, we learn the consistency knowledge by minimizing the KL-divergence with a prior Gaussian distribution.

\begin{figure}[t]
\centering
\includegraphics[width=1.0\textwidth]{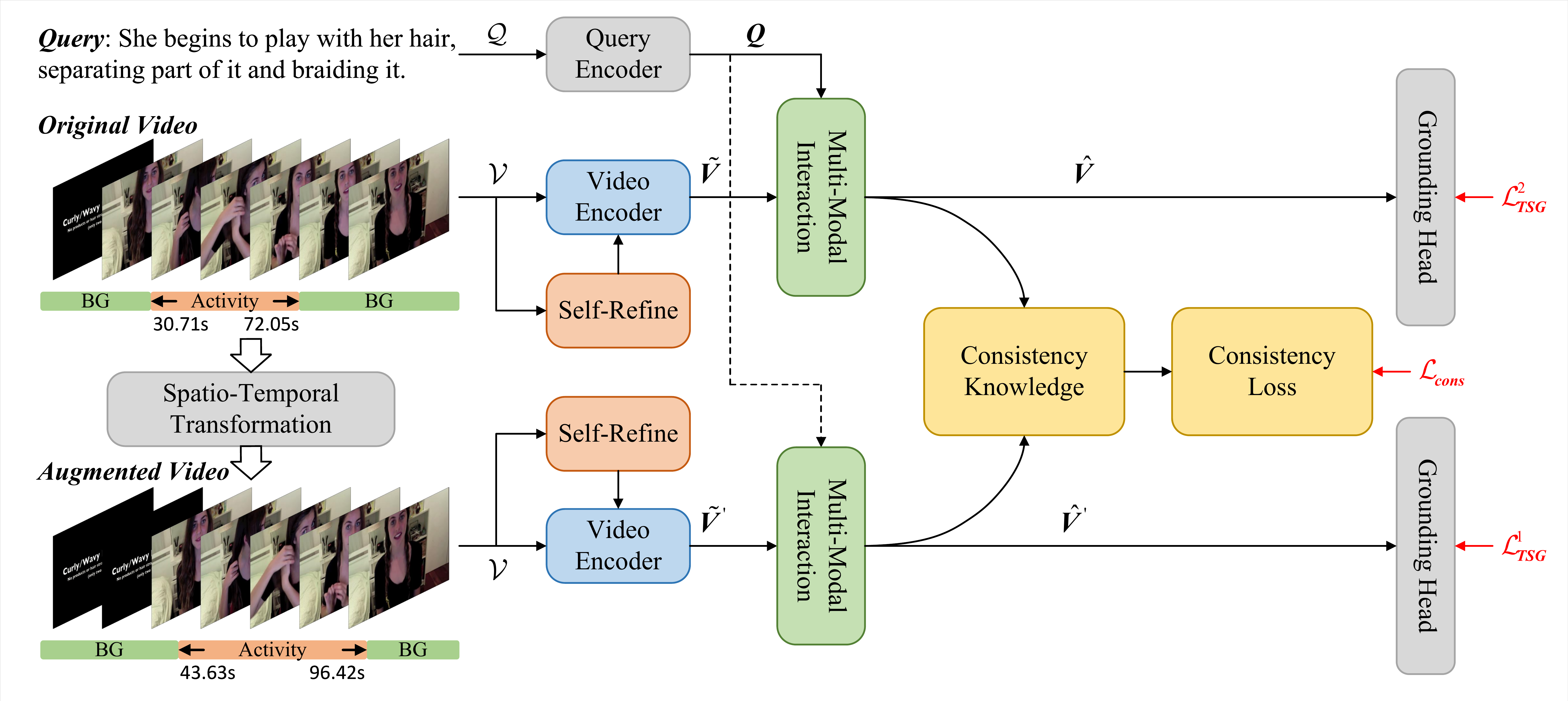}
\caption{Overall pipeline of the proposed ECRL. Given a pair of video and query input, we first apply spatio-temporal transformation on the original video to generate its augmented variant. Then, we encode both two videos with separate self-refine modules to enhance their completeness and interact them with the query. After that, we develop a self-supervised consistency learning module to discriminate the invariant query-relevant and -irrelevant frame-wise representations between two videos. At last, a shared grounding head is utilized to predict the transform-equivariant query-guided segment boundaries on them.}
\label{fig:pipeline}
\end{figure}

\section{Method}
\subsection{Problem Definition and Overview}
Given an untrimmed video $\mathcal{V}$ and a sentence query $\mathcal{Q}$, we represent the video as $\mathcal{V}=\{v_t\}^{T}_{t=1}$ frame
\footnote{In this paper, the frame is a general concept for an actual video frame or a video clip which consists of a few consecutive frames.}
-by-frame, where $v_t$ is the $t$-th frame and $T$ is the number of total frames. Similarly, the query with $N$ words is denoted as $\mathcal{Q}=\{q_n\}^{N}_{n=1}$ word-by-word. The TSG task aims to localize the start and end timestamps $(\tau_s, \tau_e)$ of a specific segment in video $\mathcal{V}$, which refers to the corresponding semantic of query $\mathcal{Q}$.
However, previous works not only rely on large amount of video-query pairs for training, but also tend to easily fit the data distribution bias during the model learning.

Therefore, we propose a novel Equivariant Consistency Regulation Learning (ECRL) framework to alleviate the above issues in a self-supervised manner.
As shown in Figure~\ref{fig:pipeline}, we first introduce a  spatio-temporal transformation module to apply various video-level augmentations on each video to synthesize new video samples for assisting the model training. Considering the adjacent frames in augmented video tend to be discrete and incomplete, we further devise a self-refine module to enhance their smoothness.
Note that, the new video samples not only can enrich the training data, but also can serve as contrastive samples for improving the model generalization ability. 
Therefore, we then present a novel self-supervised consistency loss (SSCL) to discriminate the frame-wise representations between both original and augmented videos for capturing their consistent
query-related semantics.
At last, a shared grounding head is utilized to predict the transform-equivariant query-guided segment boundaries on both original and augmented videos. We illustrate the details of each component in the following.

\subsection{Spatio-Temporal Transformation}
We first introduce the detailed spatio-temporal transformation step of our method to construct the augmented samples for better assisting the model learning. This data augmentation process is crucial to avoid trivial solutions in self-supervised learning \cite{chen2020simple}. Different from prior methods designed for image data which only require spatial augmentations, we introduce a series of spatio-temporal data augmentations to further increase the variety of videos.

\noindent \textbf{Temporal transformation.}
For temporal data augmentation, since each video contains one specific segment corresponding to the sentence query, we first split each video into three parts (sub-video $\mathcal{V}_{left}$ before the target segment, sub-video $\mathcal{V}_{seg}$ of the target segment, and sub-video $\mathcal{V}_{right}$ after the target segment) and then perform temporal transformations on them, respectively.
In detail, similar to the resize operation in image processing, we randomly perform up-sampling or down-sampling on each sub-video with different sampling ratio $r_{left},r_{seg},r_{right} \in [1-\alpha,\alpha]$ along the time dimension uniformly for changing their lengths.
Particularly, as for the sub-video with 0 frame, empty frames are padded on it before applying the sampling operation.
At last, we compose the three augmented sub-videos into a joint long video $\mathcal{V}'$, and uniformly \textit{sample} it to the same length $T$ as $\mathcal{V}$ to generate the final augmented video.

\noindent \textbf{Spatial transformation.}
For spatial data augmentation, we directly apply several spatial data augmentations, including random crop and resize, random color distortions, and random Gaussian blur, on video $\mathcal{V}'$.

\subsection{Multi-Model Encoding and Interaction}
\noindent \textbf{Video encoding.}
For the original video $\mathcal{V}$ and its augmented sample $\mathcal{V}'$, we first extract their frame-wise features by a pre-trained C3D network \cite{tran2015learning} as $\bm{V}=\{\bm{v}_t\}_{t=1}^{T}, \bm{V}'=\{\bm{v}_t'\}_{t=1}^{T} \in \mathbb{R}^{T \times D}$, where $D$ is the feature dimension.
Considering the sampled video sequence $\bm{V}'$ tends to be discrete and incomplete, we then introduce a \textbf{self-refine module} to utilize both \textit{temporal} and \textit{semantic} context information to smooth the consecutive frames. Specifically, we first construct
a fully connected graph over $\bm{V}'$ where each node is a single frame. Let $\bm{E} \in \mathbb{R}^{T\times T}$ be the adjacency matrix of graph, $\bm{E}_{i,j}$ is the edge weight between node $i$ and $j$. Intuitively, temporally neighboring frames are more likely to have correlated content. Therefore, we define the \textit{temporal} adjacency weight as follows:
\begin{equation}
    \bm{E}_{i,j}^{tem} = e^{-\frac{|i-j|^2}{2\sigma^2}},
\end{equation}
where $\sigma$ is empirically set as $5$ in all experiments.
For the \textit{semantic} similarity of frames $i$ and $j$, we directly evaluate it by measuring their cosine similarity as follows:
\begin{equation}
    \bm{E}_{i,j}^{sem} = cos(\bm{v}_i',\bm{v}_j')= \frac{\bm{v}_i' (\bm{v}_j')^{\top}}{\parallel \bm{v}_i' \parallel_2 \parallel \bm{v}_j' \parallel_2 }.
\end{equation}
The final inter-node similarity $\bm{E}_{i,j}$ is calculated by element-wise multiplication as $\bm{E}_{i,j} = \bm{E}_{i,j}^{tem} \cdot \bm{E}_{i,j}^{sem}$. After that, we iteratively update and refine $\bm{V}'$ as follows:
\begin{equation}
    \widetilde{\bm{v}_i}' = \sum_{j=1}^T \bm{E}_{i,j} \bm{v}_j', \quad \widetilde{\bm{v}_i}' \in \widetilde{\bm{V}}'
\end{equation}
where we empirically find that an overall iteration of 3 times will strike a good balance of accuracy and computational complexity.
We can also enrich the self-contexts of $\bm{V}$ in the same manner and denote its final feature as $\widetilde{\bm{V}}$.
At last, we employ a self-attention \cite{vaswani2017attention} layer and a BiLSTM \cite{Mike1997} to capture the long-range dependencies within each video. 

\noindent \textbf{Query encoding.}
For the sentence query $\mathcal{Q}$, we first generate the word-level features by using the Glove embedding \cite{pennington2014glove}, and then also employ a self-attention layer and a BiLSTM layer to further encode the query features as $\bm{Q}=\{\bm{q}_n\}_{n=1}^{N} \in \mathbb{R}^{N \times D}$.

\noindent \textbf{Multi-modal interaction.}
After obtaining the encoded features $\widetilde{\bm{V}}',\widetilde{\bm{V}},\bm{Q}$, we utilize a co-attention mechanism \cite{lu2016hierarchical} to capture the cross-modal interactions between video and query features. Specifically, for pair $(\widetilde{\bm{V}}',\bm{Q})$, we first calculate the similarity scores between $\widetilde{\bm{V}}'$ and $\bm{Q}$ as:
\begin{equation}
    \bm{S} = \widetilde{\bm{V}}'(\bm{Q}\bm{W}_S)^{\top} \in \mathbb{R}^{T\times N},
\end{equation}
where $\bm{W}_S \in \mathbb{R}^{D\times D}$ projects the query features into the same latent space as the video. Then, we compute two attention weights as:
\begin{equation}
    \bm{A} = \bm{S}_r (\bm{Q}\bm{W}_S) \in \mathbb{R}^{T\times D}, \bm{B} = \bm{S}_r \bm{S}_c^{\top} \widetilde{\bm{V}}' \in \mathbb{R}^{T\times D},
\end{equation}
where $\bm{S}_r$ and $\bm{S}_c$ are the row- and column-wise softmax results of $\bm{S}$, respectively. We compose the final query-guided video representation by learning its sequential features as:
\begin{equation}
    \widehat{\bm{V}} = BiLSTM([\widetilde{\bm{V}}';\bm{A};\widetilde{\bm{V}}'\odot \bm{A};\widetilde{\bm{V}}'\odot \bm{B}]) \in \mathbb{R}^{T\times D},
\end{equation}
where $\widehat{\bm{V}}'=\{\widehat{\bm{v}}_t'\}^{T}_{t=1}$, $BiLSTM(\cdot)$ denotes the BiLSTM layers, $[;]$ is the concatenate operation, and $\odot$ is the element-wise multiplication. In the same way, we can generate another query-guided video features $\widehat{\bm{V}}$ from the pair $(\widetilde{\bm{V}},\bm{Q})$.

\subsection{Transform-Equivariant Consistency Learning}
Human visual perception shows good consistency for query-based segment localization when they watch the video at different playback rates. For example, when we watch a video that contains a specific activity, the corresponding semantic of the video segment will not change, yet the duration and temporal boundary of the segment will change as the playback rate varies. State differently, the query-related activity of the video is invariant to different playback rates, while its temporal boundary is equivariant. Therefore, to make our model have the equivariant property between the augmented video and its original one, we propose transform-equivariant consistency learning to maximize their agreements.

\begin{figure}[t]
\centering
\includegraphics[width=0.7\textwidth]{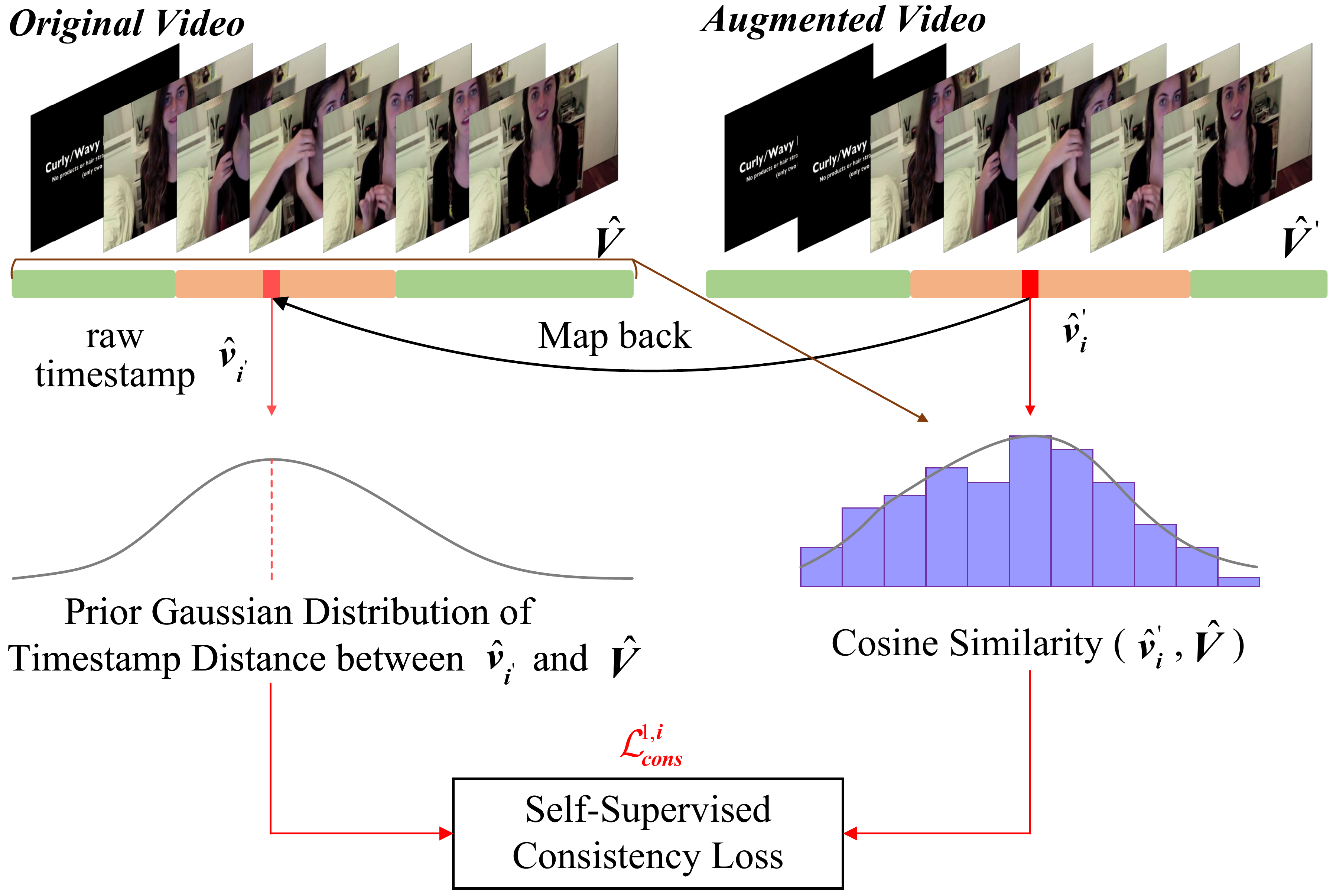}
\caption{Illustration of the proposed consistency loss.
For $\widehat{\bm{v}}_i'$ in $\widehat{\bm{V}}'$, we first compute a prior Gaussian distribution of timestamp distance between the raw timestamp $i'$ of $\widehat{\bm{v}}_i'$ and all timestamps in $\widehat{\bm{V}}$ conditioned on the sub-video correspondence. Then the semantic similarity distribution between $\widehat{\bm{v}}_i'$ and $\widehat{\bm{V}}$ is calculated, and we minimize the KL-divergence of these two distributions for discriminating the frame-wise representations.
}
\label{fig:consistency}
\end{figure}

\noindent \textbf{How to learn the consistency knowledge?} For each video, the consistency knowledge denotes that the feature of the original frame in $\mathcal{V}_{left},\mathcal{V}_{seg},\mathcal{V}_{right}$ should be semantically-invariant to the feature of the augmented frame in the same $\mathcal{V}_{left},\mathcal{V}_{seg},\mathcal{V}_{right}$, respectively. In this way, the model is able to discriminate the semantics of $\mathcal{V}_{left},\mathcal{V}_{seg},\mathcal{V}_{right}$ with different playback rate, thus predicting the equivariant segment in the augmented video.
To discriminate and learn the invariant frame-wise representations between two videos, a general idea is to take each corresponding frame as reference frame and take the other frames as negative ones.
However, videos provide abundant sequential information, and the neighboring frames around the reference frame are highly correlated. Thus, directly regarding these frames (especially the segment boundaries) as negatives may hurt the representation learning. 
To alleviate this issue, we present a novel self-supervised consistency loss (SSCL), which optimizes the frame-wise features by minimizing the Kullback–Leibler (KL) divergence \cite{goldberger2003efficient} between the sequence similarity of two videos and a prior Gaussian distribution \cite{piergiovanni2019temporal}, to capture the consistency knowledge.
As shown in Figure~\ref{fig:consistency}, to discriminate a single frame in $\widehat{\bm{V}}'$ with the entire video $\widehat{\bm{V}}$, we first compute a prior Gaussian distribution of their timestamp distance in the original video, and then calculate the semantic similarity distribution between each single augmented frame and the entire original video. At last, we minimize the KL-divergence of the similarity distribution and the Gaussian distribution in the feature space.

\noindent \textbf{Formulation of the consistency loss.}
Specifically, given the $i$-th augmented frame $\widehat{\bm{v}}_i'$ in $\widehat{\bm{V}}'$, we first find its raw video timestamp $i'$ in original video $\mathcal{V}$. Since video $\widehat{\bm{V}}$ is extracted from the original video, its timestamps $\{1,2,...,T\}$ are already the raw ones.
Due to the fact that temporally adjacent frames are more highly correlated than those far away ones, we assume the similarity between $\widehat{\bm{v}}_i'$ and $\widehat{\bm{V}}$ should follow a prior Gaussian distribution of timestamp distance between $i'$ and $\{1,2,...,T\}$. Let $G(x)=\frac{1}{\sigma \sqrt{2\pi}} e^{-\frac{x^2}{2\sigma^2}}$ denotes the Gaussian function, we use the KL-divergence to formulate the loss of $i$-th augmented frame in $\widehat{\bm{V}}'$ as follows:
\begin{equation}
    \mathcal{L}_{cons}^{1,i} = - \sum_j^T w_{ij} log \frac{e^{cos(\widehat{\bm{v}}_i',\widehat{\bm{v}}_j)}}{\sum_{t=1}^T e^{cos(\widehat{\bm{v}}_i',\widehat{\bm{v}}_t)}},
\end{equation}
\begin{equation}
    w_{ij} = \frac{G(i'-j)}{\sum_{t=1}^T G(i'-t)},
\end{equation}
where $w_{ij}$ is the normalized Gaussian weight. 
\textit{Note that}, for each frame in a sub-video, we apply an additional down-weight 0.5 to the Gaussian value of the frames in other sub-videos for better distinguishing.
Similarly, we can calculate the loss of $i$-th original frame $\mathcal{L}_{cons}^{2,i}$ for $\widehat{\bm{V}}$. Therefore, the overall SSCL loss function is formulated as:
\begin{equation}
    \mathcal{L}_{cons} = \frac{1}{T} \sum_{i=1}^T (\mathcal{L}_{cons}^{1,i}+\mathcal{L}_{cons}^{2,i}).
\end{equation}

\subsection{Grounding Heads}
To predict the target segments with the features $\widehat{\bm{V}}',\widehat{\bm{V}}$ for both augmented and original videos, we employ the efficient proposal-free prediction head to regress the start and end timestamps of the segment. Specifically, for video $\widehat{\bm{V}}'$, we utilize two separate LSTM layers to successively predict the start and end scores on each video frame as:
\begin{equation}
\label{score1}
    \bm{h}^s_t = LSTM_{start}(\widehat{\bm{v}}_t',\bm{h}^s_{t-1}), \ C^s_t = [\widehat{\bm{v}}_t';\bm{h}^s_t] \bm{W}_s + \bm{b}_s,
\end{equation}
\begin{equation}
\label{score2}
    \bm{h}^e_t = LSTM_{end}(\widehat{\bm{v}}_t',\bm{h}^e_{t-1}), \ C^e_t = [\widehat{\bm{v}}_t';\bm{h}^e_t] \bm{W}_e + \bm{b}_e,
\end{equation}
where $\bm{h}$ is the hidden state of LSTM layer, $C^s_t,C^e_t$ denote the scores of start and end boundaries at $t$-th frame. We utilize the cross-entropy loss function $\mathcal{L}_{ce}$ to supervise the grounding on the augmented video as:
\begin{equation}
    \mathcal{L}_{TSG}^1 = \frac{1}{2T} \sum_{t=1}^T [\mathcal{L}_{ce}(C^s_t,\widehat{C}^s_t)+\mathcal{L}_{ce}(C^e_t,\widehat{C}^e_t)],
\end{equation}
where $\widehat{C}^s_t,\widehat{C}^e_t$ are the ground-truth labels. Similarly, we can formulate the loss function $\mathcal{L}_{TSG}^2$ on the original video $\widehat{\bm{V}}$.

\noindent \textbf{Training.} During the training, we jointly optimize two grounding losses of two videos and the consistency loss as:
\begin{equation}
    \mathcal{L}_{overall} = \mathcal{L}_{TSG}^1 + \mathcal{L}_{TSG}^2 + \lambda \mathcal{L}_{cons},
\end{equation}
where $\lambda$ is the balanced weight.

\noindent \textbf{Testing.}
During the inference, we directly construct the top-n segments by considering the summed scores of the selected start and end boundary timestamps for each video.

\section{Experiments}
\subsection{Dataset and Evaluation}
\noindent \textbf{ActivityNet.}
ActivityNet \cite{krishna2017dense} contains 20000 untrimmed videos with 100000 descriptions from YouTube. Following public split, we use 37417, 17505, and 17031 sentence-video pairs for training, validation, testing.

\noindent \textbf{TACoS.}
TACoS \cite{regneri2013grounding} is widely used on TSG task and contain 127 videos. We use the same split as \cite{gao2017tall}, which includes 10146, 4589, 4083 query-segment pairs for training, validation and testing.

\noindent \textbf{Charades-STA.}
Charades-STA is built on \cite{sigurdsson2016hollywood}, which focuses on indoor activities. 
As Charades dataset only provides video-level paragraph description, the temporal annotations of Charades-STA are generated in a semi-automatic way.
In total,  are 12408 and 3720 moment-query pairs in the training and testing sets.

\noindent \textbf{Evaluation.} 
Following previous works \cite{gao2017tall,yang2022video}, we adopt ``R@n, IoU=m" as our evaluation metric, which is defined as the percentage of at least one of top-n selected moments having IoU larger than m.

\subsection{Implementation Details}
We implement our model in PyTorch. To extract video features, we utilize pre-trained C3D \cite{tran2015learning} to encode each video frames on ActivityNet, TACoS, and utilize pre-trained I3D \cite{carreira2017quo} on Charades-STA. Since some videos are overlong, we pre-set the length $T$ of video feature sequences to 200 for ActivityNet and TACoS datasets, 64 for Charades-STA dataset. As for sentence encoding, we set the length of word feature sequences to 20, and utilize Glove embedding \cite{pennington2014glove} to embed each word to 300 dimension features. The hidden state dimension of BiLSTM networks is set to 512. The dimension $D$ is set to 1024, and the weight $\lambda$ is set to 5.0. During the video spatio-temporal transformation, we set the parameter $\alpha=0.8$.
During the training, we use an Adam optimizer with the leaning rate of 0.0001. The model is trained for 100 epochs to guarantee its convergence with a batch size of 64 (128 samples). All the experiments are implemented on a single NVIDIA TITAN XP GPU.

\begin{table}[t!]
    \small
    \centering
    \caption{Performance compared with the state-of-the-arts TSG methods on ActivityNet, TACoS, and Charades-STA datasets.}
    \setlength{\tabcolsep}{0.2mm}{
    \begin{tabular}{c|cccc|cccc|cccc}
    \hline \hline
    \multirow{3}*{Method} & \multicolumn{4}{c|}{ActivityNet} & \multicolumn{4}{c|}{TACoS} & \multicolumn{4}{c}{Charades-STA} \\ \cline{2-5} \cline{6-9} \cline{10-13}
    ~ & R@1, & R@1, & R@5, & R@5, & R@1, & R@1, & R@5, & R@5, & R@1, & R@1, & R@5, & R@5, \\ 
    ~ & IoU=0.5 & IoU=0.7 & IoU=0.5 & IoU=0.7 & IoU=0.3 & IoU=0.5 & IoU=0.3 & IoU=0.5 & IoU=0.5 & IoU=0.7 & IoU=0.5 & IoU=0.7 \\ \hline
    TGN & 28.47 & - & 43.33 & - & 21.77 & 18.90 & 39.06 & 31.02 & - & - & - & -\\
    CBP & 35.76 & 17.80 & 65.89 & 46.20 & 27.31 & 24.79 & 43.64 & 37.40 & 36.80 & 18.87 & 70.94 & 50.19 \\
    SCDM & 36.75 & 19.86 & 64.99 & 41.53 & 26.11 & 21.17 & 40.16 & 32.18 & 54.44 & 33.43 & 74.43 & 58.08 \\
    LGI & 41.51 & 23.07 & - & - & - & - & - & - & 59.46 & 35.48 & - & - \\
    BPNet & 42.07 & 24.69 & - & - & 25.96 & 20.96 & - & - & 50.75 & 31.64 & - & - \\
    VSLNet & 43.22 & 26.16 & - & - & 29.61 & 24.27 & - & - & 54.19 & 35.22 & - & - \\
    CMIN & 43.40 & 23.88 & 67.95 & 50.73 & 24.64 & 18.05 & 38.46 & 27.02 & - & - & - & - \\
    IVG-DCL & 43.84 & 27.10 & - & - & 38.84 & 29.07 & - & - & 50.24 & 32.88 & - & - \\
    2DTAN & 44.51 & 26.54 & 77.13 & 61.96 & 37.29 & 25.32 & 57.81 & 45.04 & 39.81 & 23.25 & 79.33 & 51.15 \\
    DRN & 45.45 & 24.36 & 77.97 & 50.30 & - & 23.17 & - & 33.36 & 53.09 & 31.75 & 89.06 & 60.05 \\
    CBLN & 48.12 & 27.60 & 79.32 & 63.41 & 38.98 & 27.65 & 59.96 & 46.24 & 61.13 & 38.22 & 90.33 & 61.69 \\ 
    MMN & 48.59 & 29.26 & 79.50 & 64.76 & 39.24 & 26.17 & 62.03 & 47.39 & 47.31 & 27.28 & 83.74 & 58.41 \\ 
    MGSL & 51.87 & 31.42 & 82.60 & 66.71 & 42.54 & 32.27 & 63.39 & 50.13 & 63.98 & 41.03 & 93.21 & 63.85 \\
    \hline
    \textbf{ECRL} & \textbf{54.24} & \textbf{32.98} & \textbf{84.77} & \textbf{68.35} & \textbf{45.20} & \textbf{34.43} & \textbf{65.74} & \textbf{51.86} & \textbf{65.37} & \textbf{42.69} & \textbf{94.52} & \textbf{65.18} \\ \hline
    \end{tabular}}
    \label{tab:compare}
\end{table}

\subsection{Comparison with State-of-the-Arts}
\noindent \textbf{Compared methods.}
We compare the proposed ECRL with
state-of-the-art TSG methods on three datasets: TGN \cite{chen2018temporally}, CBP \cite{wang2019temporally}, SCDM \cite{yuan2019semantic}, BPNet \cite{xiao2021boundary}, CMIN \cite{zhang2019cross}, 2DTAN \cite{zhang2019learning}, DRN \cite{zeng2020dense}, CBLN \cite{liu2021context}, MMN \cite{wang2022negative}, and MGSL \cite{liu2022memory}, LGI \cite{mun2020local}, VSLNet \cite{zhang2020span}, IVG-DCL \cite{nan2021interventional}.

\noindent \textbf{Quantitative comparison.}
As shown in Table~\ref{tab:compare}, we compare our proposed ECRL model with the existing TSG methods on three datasets, where our ECRL outperforms all the existing methods across different criteria by a large margin. Specifically, on ActivityNet dataset, compared to the previous best proposal-based method MGSL, we do not rely on large numbers of pre-defined proposals and outperform it by 2.37\%, 1.56\%, 2.17\%, 1.64\% in all metrics, respectively. Compared to the previous best bottom-up method IVG-DCL, our ECRL brings significant improvement of 10.40\% and 5.88\% in the R@1, IoU=0.5 and R@1, IoU=0.7 metrics.
On TACoS dataset, the cooking activities take place in the same kitchen scene with some slightly varied cooking objects, thus it is hard to localize such fine-grained activities. Compared to the top ranked method MGSL, our model still achieves the best results on the strict metrics R@1, IoU=0.5 and R@5, IoU=0.5 by boosting 2.16\% and 1.73\%, which validates that ECRL is able to localize the segment boundary more precisely. 
On Charades-STA dataset, we outperform the MGSL by 1.39\%, 1.66\%, 1.31\% and 1.33\% in all metrics, respectively. The main reasons for our proposed model outperforming the competing models lie in two folds:
1) Our newly proposed self-supervised consistency loss learns more accurate frame-wise representations for alleviating the data distribution bias, improving the generalization-ability of the model. 2) The augmented video helps the model capture the invariant query-related semantics for better predicting the transform-equivariant segment boundaries.

We further compare our method with existing works \cite{lan2022closer,zhang2021towards} on the de-biased datasets \cite{lan2022closer}, \textit{i.e.}, ActivityNet-CD and Charades-CD. As shown in Table~\ref{tab:compare_cd}, our method still outperforms the other methods by a large margin.

\begin{table}[t!]
    \small
    \centering
    \caption{Performance comparison on ActivityNet-CD and Charades-CD datasets \cite{lan2022closer}.}
    \setlength{\tabcolsep}{2.0mm}{
    \begin{tabular}{c|cc|cc}
    \hline \hline
    \multirow{2}*{Method} & \multicolumn{2}{c|}{ActivityNet-CD} & \multicolumn{2}{c}{Charades-CD} \\ \cline{2-3} \cline{4-5} 
    ~ & test-IID & test-OOD & test-IID & test-OOD \\ \hline
    Zhang \textit{et al.} \cite{zhang2021towards} & 28.11 & 14.67 & 38.87 & 32.70 \\
    Lan \textit{et al.} \cite{lan2022closer} & 31.44 & 11.66 & 34.71 & 22.70
    \\
    \hline
    \textbf{ECRL} & \textbf{34.37} & \textbf{19.95} & \textbf{41.59} & \textbf{34.98}\\ \hline
    \end{tabular}}
    \label{tab:compare_cd}
\end{table}

\begin{table}[t!]
    \small
    \centering
    \caption{Efficiency comparison in terms of second per video (SPV) and parameters (Para.), where our method ECRL is much efficient with relatively lower model size.}
    \setlength{\tabcolsep}{1.6mm}{
    \begin{tabular}{c|ccccccc}
    \hline \hline 
    ~ & ACRN & CTRL & TGN & 2DTAN & MGSL & VLSNet & \textbf{ECRL} \\ \hline
    SPV $\downarrow$ & 4.31 & 2.23 & 0.92 & 0.57 & 0.10 & 0.07 & \textbf{0.06} \\ \hline
    Para. $\downarrow$ & 128 & \textbf{22} & 166 & 232 & 203 & 48 & 54 \\ \hline
    \end{tabular}}
    \label{tab:efficient}
\end{table}

\noindent \textbf{Comparison on efficiency.}
We evaluate the efficiency of our proposed ECRL model, by fairly comparing its running time and model size in inference phase with existing methods on a single Nvidia TITAN XP GPU on TACoS dataset. As shown in Table~\ref{tab:efficient}, it can be observed that we achieve much faster processing speeds with relatively less learnable parameters. This attributes to:
1) The proposal-based methods (ACRN, CTRL, TGN, 2DTAN, DRN) suffer from the time-consuming process of proposal generation and proposal matching. Compared to them, our grounding head is proposal-free, which is much more efficient and has less parameters.
2) The proposal-free method VLSNet utilizes convolution operation to discriminate foreground-background frames. Instead, we propose an efficient and effective consistency loss function to learn frame-wise representation.

\subsection{Ablation Study}
We perform multiple experiments to analyze different components of our ECRL framework. Unless otherwise specified, experiments are conducted on the ActivityNet dataset.

\noindent \textbf{Main ablation.}
To demonstrate the effectiveness of each component in our ECRL, we conduct ablation studies regarding the components (\textit{i.e.}, two grounding heads on augmented video and original video in backbone model, two consistency constraints in the SSCL, and the self-refine module in the video encoder) of ECRL, and show the corresponding experimental results in Table~\ref{tab:ablation1}.
In particular, the first line represents the performance of
the baseline model ($\mathcal{L}_{TSG}^2$), which only train the original video-query pairs without augmented samples and consistency loss, achieving 43.19\% and 26.72\% in R@1, IoU=0.5 and R@1, IoU=0.7.
Comparing the results in other lines of this table, we have the following observations:
1) The spatio-temporal transformation strategy constructs the augmented samples to assist the model training, which promotes the model performance (refer to line 1-2 of the table). However, the improvement is small as the consistency between augmented and original video still is not captured.
2) Both two types of consistency losses contributes a lot to the grounding performance. Specifically, each type of them is able to learn the invariant semantics between two videos for enhancing the frame-wise representation learning, thus improving performance of the transform-equivalent segment prediction (refer to line 3-4 of the table). Utilizing both of them (refer to line 5) can further boost the performance.
3) The self-refine module also contributes to the final performance (refer to line 6) by enhancing the completeness and smoothness of the discrete augmented video frames.
In total, results demonstrate the effectiveness of our each component.

\begin{table}[t!]
    \small
    \centering
    \caption{Main ablation study on ActivityNet dataset. Here, $\mathcal{L}_{TSG}^1,\mathcal{L}_{TSG}^2$ and $\mathcal{L}_{cons}^1,\mathcal{L}_{cons}^2$ are the grounding backbones and the consistency learning modules on the augmented and original video inputs, respectively. `SR' denotes the self-refine module.}
    \setlength{\tabcolsep}{2.0mm}{
    \begin{tabular}{ccccccc|cc}
    \hline \hline 
    \multicolumn{2}{c}{Backbone} & & \multicolumn{2}{c}{Consistency} & & \multirow{2}*{SR} & R@1, & R@1, \\ \cline{1-2} \cline{4-5}
    $\mathcal{L}_{TSG}^1$ & $\mathcal{L}_{TSG}^2$ & & $\mathcal{L}_{cons}^1$ & $\mathcal{L}_{cons}^2$ & & ~ & IoU=0.5 & IoU=0.7 \\ \hline
    $\times$ & $\checkmark$ & & $\times$ & $\times$ & & $\times$ & 43.19 & 26.72
    \\
    $\checkmark$ & $\checkmark$ & & $\times$ & $\times$ & & $\times$ & 44.33 & 27.46
    \\
    $\checkmark$ & $\checkmark$ & & $\times$ & $\checkmark$ & & $\times$ & 49.28 & 30.01
    \\
    $\checkmark$ & $\checkmark$ & & $\checkmark$ & $\times$ & & $\times$ & 49.84 & 30.47
    \\
    $\checkmark$ & $\checkmark$ & & $\checkmark$ & $\checkmark$ & & $\times$ & 51.79 & 31.65
    \\ \hline
    $\checkmark$ & $\checkmark$ & & $\checkmark$ & $\checkmark$ & & $\checkmark$ & \textbf{54.24} & \textbf{32.98}
    \\ \hline
    \end{tabular}}
    \label{tab:ablation1}
\end{table}

\begin{table}[t!]
    \small
    \centering
    \caption{Effect of the consistency loss on ActivityNet.}
    \setlength{\tabcolsep}{2.0mm}{
    \begin{tabular}{c|c|cc}
    \hline \hline
    \multirow{2}*{Components} & \multirow{2}*{Changes} & R@1, & R@1,\\
    ~ & ~ & IoU=0.5 & IoU=0.7 \\ \hline
    \multirow{5}*{\tabincell{c}{Consistency\\ Knowledge}} & w/ Gaussian prior & \textbf{54.24} & \textbf{32.98}  \\
    ~ & w/o Gaussian prior & 50.18 & 30.06\\  \cline{2-4}
    ~ & $\sigma^2$=1 & 52.47 & 31.29\\ 
    ~ & $\sigma^2$=25 & \textbf{54.24} & \textbf{32.98} \\
    ~ & $\sigma^2$=100 & 51.65 & 30.83 \\ 
    \hline
    \multirow{4}*{\tabincell{c}{Consistency\\ Loss}} & w/ only $\mathcal{V}_{seg}$ & 52.38 & 31.46 \\
    ~ & w/ $\mathcal{V}_{seg},\mathcal{V}_{right}$ & 53.59 & 32.29 \\ 
    ~ & w/ $\mathcal{V}_{left},\mathcal{V}_{seg}$ & 53.64 & 32.31 \\ 
    ~ & w/ $\mathcal{V}_{left},\mathcal{V}_{seg},\mathcal{V}_{right}$ & \textbf{54.24} & \textbf{32.98} \\
    \hline
    \end{tabular}}
    \label{tab:ablation2}
\end{table}

\noindent \textbf{Effect of the proposed consistency loss.}
As shown in Table~\ref{tab:ablation2}, we investigate the effectiveness of the proposed self-supervised consistency module. We have the following observations from these results:
1) As for consistency knowledge, directly regarding the frames as negative samples (w/o Gaussian prior) achieves worse performance than the w/ Gaussian prior variant. It indicates that the neighboring frames around the reference frame are highly correlate, a more soft gaussian-based contrastive way can lead more fine-grained representation learning in videos.
Beside, the variance $\sigma^2$ of the prior Gaussian distribution controls how the adjacent frames are similar to the reference frame, on the assumption. It shows that too small variance ($\sigma^2=1$) or too large variance ($\sigma^2=100$) degrades the performance. We use $\sigma^2=25$ by default.
2) As for consistency loss, we find that discriminating the frames in $\mathcal{V}_{seg}$ already achieves very great grounding performance by learning the query-invariant semantics. Discriminating the frames in $\mathcal{V}_{left},\mathcal{V}_{right}$ can further boost the performance by distinguishing foreground-background.

\noindent \textbf{Study on different temporal transformations.}
Here, we study the different temporal transformation, including transforming which sub-video and the sampling ratio $\alpha$. Table~\ref{tab:ablation3} shows the results. From the table, we can see that all three sub-videos are crucial for discriminating the query-relevant and query-irrelevant frame representations. Further, the whole model achieves the best performance when the sampling ratio $\alpha$ is set to 0.8.

\begin{table}[t!]
    \small
    \centering
    \caption{Study on temporal transformation on ActivityNet.}
    \setlength{\tabcolsep}{2.0mm}{
    \begin{tabular}{c|c|cc}
    \hline \hline
    \multirow{2}*{Components} & \multirow{2}*{Changes} & R@1, & R@1,\\
    ~ & ~ & IoU=0.5 & IoU=0.7 \\ \hline
    \multirow{3}*{\tabincell{c}{Transform\\ where?}} & only $\mathcal{V}_{seg}$ & 51.96 & 31.35 \\
    ~ & $\mathcal{V}_{left},\mathcal{V}_{right}$ & 50.78 & 30.42 \\
    ~ & $\mathcal{V}_{left},\mathcal{V}_{seg},\mathcal{V}_{right}$ & \textbf{54.24} & \textbf{32.98} \\ \hline
    \multirow{3}*{\tabincell{c}{Hyper-\\ Parameter}} & $\alpha=0.7$ & 53.47 & 32.51 \\
    ~ & $\alpha=0.8$ & \textbf{54.24} & \textbf{32.98} \\
    ~ & $\alpha=0.9$ & 54.19 & 32.96 \\
    \hline
    \end{tabular}}
    \label{tab:ablation3}
\end{table}

\noindent \textbf{Evaluation of different backbone models.}
Our proposed self-supervised consistency learning can serve as a ``plug-and-play" for existing TSG methods. As shown in Table~\ref{tab:ablation4}, we demonstrate the effectiveness of our proposed method by directly applying our augmentation strategy and the SSCL to other TSG models. It shows that our method helps to learn more discriminate frame-wise features for learning query-invariant semantics and predicting transform-equivariant segment, improving the generalization-ability of the model.

\begin{table}[t!]
    \small
    \centering
    \caption{Evaluation of different grounding backbones. We apply our augmentation and SSCL on existing TSG models.}
    \setlength{\tabcolsep}{2.0mm}{
    \begin{tabular}{c|c|cc}
    \hline \hline
    \multirow{2}*{Methods} & \multirow{2}*{Changes} & R@1, & R@1,\\
    ~ & ~ & IoU=0.5 & IoU=0.7 \\ \hline
    \multirow{2}*{LGI} & Original & 41.51 & 23.07 \\
    ~ & + ECRL & \textbf{50.25} & \textbf{28.63} \\ \hline
    \multirow{2}*{VSLNet} & Original & 43.22 & 26.16 \\
    ~ & + ECRL & \textbf{53.48} & \textbf{33.76} \\ \hline
    \end{tabular}}
    \label{tab:ablation4}
\end{table}

\begin{figure}[t]
\centering
\includegraphics[width=0.7\textwidth]{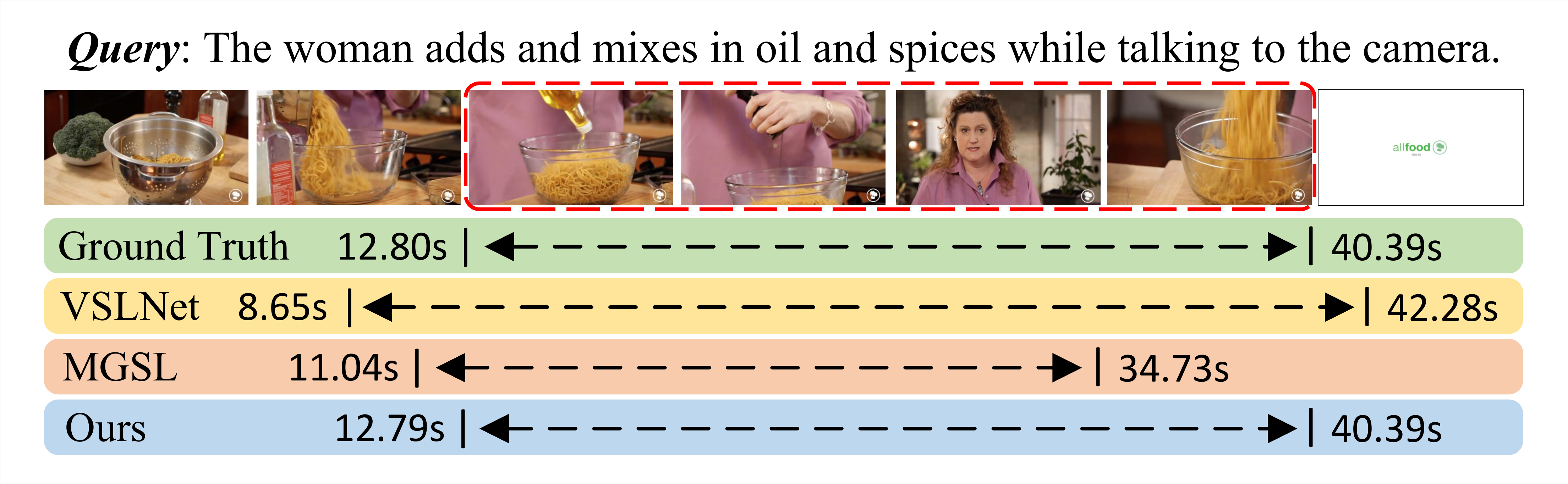}
\caption{The visualization examples of grounding results.}
\label{fig:result}
\end{figure}

\subsection{Visualization}
As shown in Figure~\ref{fig:result}, we give the qualitative examples of the grounding results. Compared to VSLNet and MGSL, our method can learn more discriminative frame-wise representations and ground the segment more accurately.

\section{Conclusion}
In this paper, we propose a novel Equivariant Consistency Regulation Learning (ECRL) framework to enhance the generalization-ability and robustness of the TSG model. Specifically, we introduce a video data augmentation strategy to construct synthetic samples, and propose a self-supervised consistency loss to learn the semantic-invariant frame-wise representations for assisting model learning and predicting transform-equivariant segment boundaries.
Experimental results on three challenging benchmarks demonstrate the effectiveness of the proposed ECRL.

\bibliographystyle{ACM-Reference-Format}
\bibliography{acmart}

\end{document}